%% file: camera_ready.tex
\documentclass[10pt,twocolumn,letterpaper]{article}

\usepackage[pagenumbers]{cvpr} 

\usepackage{graphicx}
\usepackage{amsmath}
\usepackage{amssymb}
\usepackage{booktabs}
\usepackage{enumitem}

\usepackage[pagebackref=off,breaklinks,colorlinks]{hyperref}

\usepackage[capitalize]{cleveref}
\crefname{section}{Sec.}{Secs.}
\Crefname{section}{Section}{Sections}
\Crefname{table}{Table}{Tables}
\crefname{table}{Tab.}{Tabs.}

\def\cvprPaperID{9467} 
\def\confName{CVPR}
\def\confYear{2023}

\newcommand{\formattedparagraph}[1]{\noindent \textbf{#1}}

\begin{document}

\title{Enhanced Stable View Synthesis}

\author{Nishant Jain$\thanks{Equal Contribution}$\\
Indian Institute of Technology\\
Roorkee, India\\
{\tt\small njain@cs.iitr.ac.in}
\and
Suryansh Kumar$^{*}\thanks{Corresponding Author (k.sur46@gmail.com)}$\\
ETH Z\"urich\\
Switzerland\\
{\tt\small sukumar@ethz.ch}
\and
Luc Van Gool\\
ETH Z\"urich\\
Switzerland\\
{\tt\small vangool@ethz.ch}
}
\maketitle

\begin{abstract}
We introduce an approach to enhance the novel view synthesis from images taken from a freely moving camera. The introduced approach focuses on outdoor scenes where recovering accurate geometric scaffold and camera pose is challenging, leading to inferior results using the state-of-the-art stable view synthesis (SVS) method. SVS and related methods fail for outdoor scenes primarily due to (i) over-relying on the multiview stereo (MVS) for geometric scaffold recovery and (ii) assuming COLMAP computed camera poses as the best possible estimates, despite it being well-studied that MVS 3D reconstruction accuracy is limited to scene disparity and camera-pose accuracy is sensitive to key-point correspondence selection. This work proposes a principled way to enhance novel view synthesis solutions drawing inspiration from the basics of multiple view geometry. By leveraging the complementary behavior of MVS and monocular depth, we arrive at a  better scene depth per view for nearby and far points, respectively. Moreover, our approach jointly refines camera poses with image-based rendering via multiple rotation averaging graph optimization. The recovered scene depth and the camera-pose help better view-dependent on-surface feature aggregation of the entire scene. Extensive evaluation of our approach on the popular benchmark dataset, such as Tanks and Temples, shows substantial improvement in view synthesis results compared to the prior art. For instance, our method shows \textbf{1.5 dB} of PSNR improvement on the Tank and Temples. Similar statistics are observed when tested on other benchmark datasets such as FVS, Mip-NeRF 360, and DTU.
\end{abstract}

\section{Introduction}\label{sec:intro}
Image-based rendering, popularly re-branded as view synthesis, is a long-standing problem in computer vision and graphics \cite{shum2000review, tewari2020state}. This problem aims to develop a method that allows the user to seamlessly explore the scene via rendering of the scene from a sparse set of captured images \cite{shum2000review, buehler2001unstructured, kopf2014first}. Furthermore, the rendered images must be as realistic as possible for a better user experience \cite{richter2022enhancing, riegler2020free, riegler2021stable}. Currently, among the existing approaches, Riegler and Koltun stable view synthesis (SVS) approach \cite{riegler2021stable} has shown excellent results and demonstrated photorealism in novel view synthesis, without using synthetic gaming engine 3D data, unlike \cite{richter2022enhancing}. SVS is indeed stable in rendering photorealistic images from novel viewpoints for large-scale scenes. Yet, it assumes MVS \cite{schoenberger2016sfm, schonberger2016pixelwise} based dense 3D scene reconstruction and camera poses from COLMAP \cite{schoenberger2016sfm} are correct. The off-the-shelf algorithms used for 3D data acquisition and camera poses from images are, of course, popular, and to assume these algorithms could provide favorable 3D reconstruction and camera poses is not an outlandish assumption. Nonetheless, taking a step forward, in this paper, we argue that although choices made by SVS for obtaining geometric scaffold and camera poses in the pursuit of improving view synthesis is commendable, we can do better by making mindful use of fundamentals from multiple-view geometry \cite{hartleymultiple, hartley2013rotation} and recent developments in deep-learning techniques for 3D computer vision problems.

To start with, we would like to emphasize that it is clearly unreasonable, especially in an outdoor setting, to assume that multi-view stereo (MVS) can provide accurate depth for all image pixels. It is natural that pixels with low disparity will not be reconstructed well using state-of-the-art MVS approaches \cite{schonberger2016pixelwise, schoenberger2016sfm, furukawa2015multi, wang2021patchmatchnet}. Even a precise selection of multiple view images with reasonable distance between them (assume good baseline for stereo) may not be helpful due to loss of common scene points visibility, foreshortening issue, etc. \cite{szeliski2022computer}. Such issues compel the practitioner to resort to post-processing steps for refining the MVS-based 3D geometry so that it can be helpful for rendering pipeline or neural-rendering network at train time.

\begin{figure*}[t]
    \centering
    \includegraphics[scale=0.18]{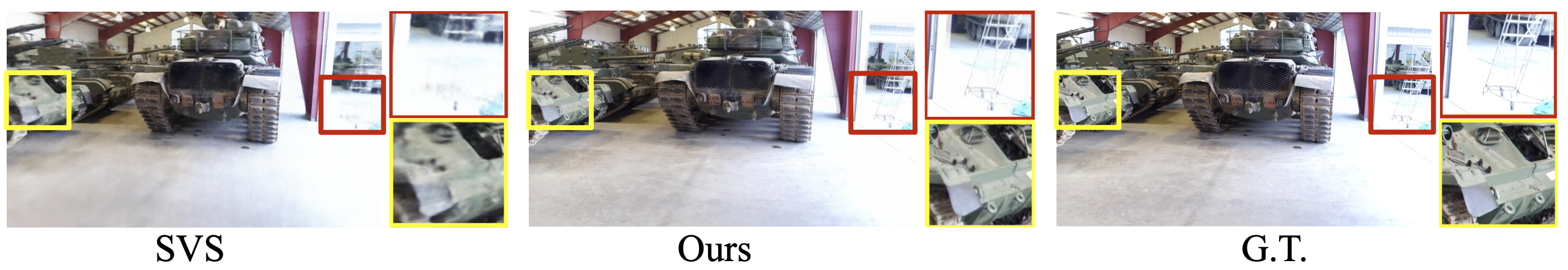}
    \caption{\textbf{Qualitative comparison.} Our result compared to the popular SVS method \cite{riegler2021stable} on the M60 scene of the tanks and temples dataset \cite{knapitsch2017tanks}. It is easy to observe that our approach can better render fine details in the scene. For this scene, the PSNR values for SVS \cite{riegler2021stable} and our method are $\mathbf{19.1}$ and $\mathbf{20.8}$, respectively, demonstrating improved PSNR result.}
    \label{fig:teaser}
\end{figure*}

Another critical component to view synthesis, which is often brushed aside in the literature is the accurate recovery of the camera poses. In neural view synthesis approaches such as \cite{riegler2021stable}, if the camera pose is wrong, the feature aggregation corresponding surface points could be misleading, providing inferior results. Therefore, we should have camera poses as accurate as possible. Unfortunately, despite the camera pose importance to this problem, discussion on improving camera pose is often ignored under the assumption that COLMAP \cite{schoenberger2016sfm} provide the best possible camera pose estimates for any possible scenarios. Practically speaking, this is generally not the case for outdoor scenes \cite{chatterjee2017robust, hartley2013rotation, Jain_2022_BMVC}. What is more surprising is that some recent benchmark datasets put COLMAP recovered poses as the ground-truth poses \cite{reizenstein2021common}. Hence, we want to get this out way upfront that a robust and better camera-pose estimates are vital for better modeling view synthesis problem.

From the above predication, it is apparent that a more mindful approach is required to make view synthesis approaches practically useful,  automatic, and valuable for real-world application. To this end, we propose a principled and systematic approach that provides a better geometric scaffold and camera poses for reliable feature aggregation of the scene's surface points, leading to improved novel-view synthesis results enabling superior photorealism.

In practice, we can have suitable initial camera poses from images using COLMAP. Yet, it must be refined further for improved image-feature aggregation corresponding to 3D surface points for neural rendering. It is well-studied in multiple-view geometry literature that we can improve and refine camera poses just from image key-point correspondences \cite{govindu2001combining, hartley2013rotation}. Accordingly, we introduce a learning-based multiple motion averaging via graph neural network for camera pose recovery, where the pose graph is initialized using COLMAP poses for refinement.

Meanwhile, it is challenging to accurately recover the 3D geometry of scene points with low or nearly-zero disparity using MVS methods \cite{szeliski2022computer, hartleymultiple}. Another bad news from the theoretical side is that a precise estimation of scene depth from a single image is unlikely\footnote{As several 3D scene points can have same image projection.},  which is a correct statement and hard to argue. The good news is that advancements in deep-learning-based monocular depth prediction have led to some outstanding results in several practical applications \cite{Ranftl2021, liuva}. Thus, at least practically, it seems possible to infer reliable monocular depth estimates up to scale. Using single image depth prediction, we can reason about the depth of scene points with low disparities. So, our proposed strategy is to use confidence based multiple-view stereo 3D that favours pixels with near-to-mid disparity and allows monocular depth estimates for the rest of the pixels. Overall depth is recovered after scaling all the scene depth appropriately using MVS reconstructed metric. 

By encoding the image features via convolutional neural networks, we map the deep features to our estimated 3D geometric scaffold of the scene. Since we have better camera poses and scene reconstruction, we obtain and aggregate accurate feature vectors corresponding to each imaging view-rays---both from the camera to the surface point and from the surface point to viewing image pixels, giving us a feature tensor. We render the new image from the features tensor via a convolutional network and simultaneously refine the camera pose. In summary, our contributions are

\begin{itemize}[leftmargin=*,topsep=0pt, noitemsep]
    \item A systematic and principled approach for improved stable view synthesis enabling enhanced photorealism.
    
    \item The introduced approach exploits the complementary nature of MVS and monocular depth estimation to recover better 3D geometric scaffold of the scene. Meanwhile, the robust camera poses are recovered using graph neural network based multiple motion averaging.
    
    \item Our approach proposes an improved loss function to jointly optimize and refine for poses, neural image rendering, and scene representation showing superior results.
\end{itemize}

\smallskip
\noindent
Our approach when tested on benchmark datasets such as Tank and Temples \cite{knapitsch2017tanks}, FVS \cite{riegler2020free}, Mip-NeRF 360\cite{barron2022mip}, and DTU \cite{jensen2014large} gives better image based rendering results with generally more than \textbf{1 dB} PSNR gain (see Fig.\ref{fig:teaser}).

\section{Background and Preliminaries}\label{sec:rel_work}
Our work integrates the best of novel view synthesis and multiple view geometry approaches in computer vision in a mindful way. Both novel view synthesis and MVS are classical problems in computer graphics and computer vision, with an exhaustive list of literature. Thus, we confine our related work discussion to the methods that is directly related to the proposed approach. The interested reader may refer to \cite{shum2000review, tewari2020state, chang2019review, ozyecsil2017survey, saputra2018visual, hartleymultiple, furukawa2015multi} for earlier and recent progress in these areas. Here, we briefly discuss relevant methods and the current state-of-the-art in neural image-based rendering.

\smallskip
\formattedparagraph{\textit{(i)} Uncalibrated Multi-View Stereo.} Given the intrinsic camera calibration matrix, we can recover camera poses and the 3D structure of the scene using two or more images \cite{furukawa2015multi, schoenberger2016sfm, schonberger2016pixelwise, kumar2017monocular, kumar2019superpixel}. One popular and easy-to-use MVS framework is COLMAP \cite{schoenberger2016sfm}, which includes several carefully crafted modules to estimate camera poses and sparse 3D structures from images. For camera pose estimation, it uses classical image key-point-based algorithms \cite{hartley1997defense, nister2004efficient, Triggs:1999:BAM:646271.685629}. As is known that such methods can provide sub-optimal solutions and may not robustly handle outliers inherent to the unstructured images. Consequently, camera poses recovered using COLMAP can be unreliable, primarily for outdoor scenes. Moreover, since the 3D reconstruction via triangulation uses sparse key points, at best accurate semi-dense 3D reconstruction of the scene could be recovered. Still, many recent state-of-the-art methods in novel view synthesis heavily rely on it \cite{mildenhall2020nerf, riegler2021stable, riegler2020free, zhang2020nerf++}.

\smallskip
\formattedparagraph{\textit{(ii)} Image-based rendering.} Earlier image-based rendering approaches enabled novel view synthesis from images without any 3D scene data under some mild assumptions about the camera, and imaging \cite{gortler1996lumigraph, levoy1996light, seitz1996view}. Later with the development of multiple-view stereo approaches and RGBD sensing modalities, several works used some form of 3D data to improve image-based rendering. Popular works along this line includes \cite{chaurasia2013depth, hedman2016scalable, buehler2001unstructured, kopf2014first, penner2017soft}. In recent years, neural network-based methods have dominated this field and enabled data-driven approach to this problem with an outstanding level of photorealism \cite{riegler2020free, riegler2021stable, zhang2020nerf++, mildenhall2021nerf, richter2022enhancing}. 

Among all the current methods we tested, SVS \cite{riegler2021stable} stands out in novel view synthesis from images taken from a freely moving camera. This paper proposes an approach closely related to the SVS  pipeline. SVS involves learning a scene representation from images using its dense 3D structure and camera poses predicted via uncalibrated MVS approaches detailed previously. The recovered 3D structure of the scene is post-processed to have a refined geometric scaffold in mesh representation. Using refined scene 3D, it learns a feature representation for a set of images and then projects it along the direction of the target view to be rendered. Finally, SVS involves re-projection from feature to image space resulting in the synthesized image for the target view. Given $\mathcal{S}$, the 3D structure for a scene, feature representation network $f_\theta$ with $\theta$ representing the network parameters, and the rendering network $\mathcal{G}_\mu$ with parameters $\mu$, the resulting image $\mathcal{I}_r$ along the pose $\textbf{p}$ is rendered as 
 
\begin{equation}
    \mathcal{I}_r = \mathcal{G}_\mu(\phi_a (\mathcal{S}, \mathcal{I}_s, \textbf{p}, f_\theta),
\end{equation}
where, $\mathcal{I}_s$ denotes the image set for a given seen. $\phi_a()$ aggregates features for all the images along a given direction in $\mathcal{I}_s$ predicted when passed through $f_\theta$.

Contrary to the choices made by the SVS, in this work, we put forward an approach for better estimation of the overall scene 3D geometric scaffold---for both low and high-disparity pixels---and camera poses. Our approach enables a better aggregation of surface features allowing an improved level of realism in novel view synthesis.

\section{Method}\label{sec:method}
We begin with a discussion about formulating and estimating a better scene representation by carefully exploiting the complementary behavior of monocular depth and multi-view stereo approaches in 3D data acquisition. During our implementation, we use confidence measures for precise reasoning of estimated 3D coming from both the modalities and their accuracy. Next, we detail integrating a camera pose-refining optimization to estimate better camera poses for accurately projecting our 3D scene representation to the target view. Finally, the proposed joint optimization objective to refine structure representation and camera poses simultaneously is discussed. 

\subsection{Overview}
Given a set of source images $\mathcal{I}_{s}$ taken from a freely moving camera, the current state-of-the-art SVS \cite{riegler2021stable} first estimates camera-poses $\mathcal{P}$ and scene 3D structure $\mathcal{S}$. Next, it encodes the images via a CNN. Then, using the estimated camera, it maps the encoded features onto the scene 3D scaffold. For each point $\mathbf{x} \in \mathbb{R}^{3}$ on the scaffold, SVS query set of images in which $\mathbf{x}$ is visible to obtain its corresponding feature vectors. This feature vector is then conditioned on the output view direction via a network to produce a new feature vector. Such a new feature vector is obtained for all the points on the scaffold to form a feature tensor, which is then decoded using CNN to synthesize the output image. Denoting the complete set of parameters for learning the feature representation and rendering as $\Theta$,  we can write SVS idea as
\begin{equation}
    \Theta \sim \Psi(\Theta|\mathcal{S},\mathcal{P},\mathcal{I}_{s}) ~\circ  \Psi(\mathcal{S},\mathcal{P}|\mathcal{I}_{s}).
\end{equation}
$\Psi(.)$ symbolizes an abstract functional. Here,  $\mathcal{S}$ and $\mathcal{P}$ are estimated using structure from motion and MVS \cite{schoenberger2016sfm,schonberger2016pixelwise}. As mentioned before, we first aim to recover a much better 3D scene recovery by utilizing the complementary nature of the monocular depth and stereo depth. Furthermore, estimate improved camera poses using neural graph-based multiple motion averaging. We denote $\mathcal{D}_s$ as monocular depth for each image in $\mathcal{I}_s$ and $\mathcal{S}_R$, $\mathcal{P}_R$ as our scene 3D  reconstruction and camera poses, respectively.  

Once we estimate $\mathcal{S}_R$, $\mathcal{P}_R$, we jointly optimize for better neural scene representation and camera poses with the neural network parameters $\Theta$ using the proposed loss function (Sec. \ref{ssec:joint_opt}). Our better scene representation and improved camera help in improved view-synthesis at test time. Our overall idea can be understood via the following equation:  
\begin{equation}
    \Theta,\mathcal{S}_R,\mathcal{P}_R \sim \Psi(\Theta,\mathcal{S}_R,\mathcal{P}_R|\mathcal{S},\mathcal{P}, \mathcal{D}_{s}, \mathcal{I}_{s})
\end{equation}

\subsubsection{3D Scene Features} \label{sec:scene_feature}

Here, we describe feature aggregation for the output view obtained using the SVS pipeline \cite{riegler2021stable} combined with the back-projected RGB-D features due to images and their monocular depth via a convolutional neural network (CNN). Given the output direction $\textbf{u}$, for each $\mathbf{x} \in \mathbb{R}^{3}$ on the geometric scaffold along $\textbf{u}$ there is a subset of source images, in which $\textbf{x}$ is visible. Assuming the total number of images in this subset to be $K$, we denote the set of view directions corresponding to these images as $\{\textbf{\text{v}}_k\}_{k = 1}^K$.

\smallskip
\formattedparagraph{\textit{(i)} SVS Features.} The source images in $\mathcal{I}_s$ are first passed through a feature extraction CNN to obtain a feature tensor $\mathcal{F}_k$ corresponding to $k^{th}$ image. Denoting $f_k(\textbf{x})$ as the feature corresponding to a point $\mathbf{x} \in \mathbb{R}^3$ in $\mathcal{F}_k$ located at the projection (or bilinear interpolation) of $\textbf{x}$ on $k^{th}$ image. To this end, SVS \cite{riegler2021stable} proposed the following aggregation function ($\phi_a$) to compute the feature for $\textbf{x}$ along $\textbf{u}$
\begin{equation}
   \phi_a(\textbf{\text{u}}, \{(\textbf{\text{v}}_k, f_k(\textbf{\text{x}}))\}) = \frac{1}{W}\sum_{k=1}^K\max(\text{0}, \text{\textbf{u}}^\text{T}\text{\textbf{v}}) f_k(\textbf{x}),
   \label{eq:agg}
\end{equation}
where, $W = \sum_{k=1}^K\max(0,{\mathbf{u}}^\text{T}\text{\textbf{v}})$ is sum of all the weights.

\smallskip
\formattedparagraph{\textit{(ii)} Monocular Features.} 
Given an image, we predict its depth with a per-pixel confidence score. For this, we use an existing monocular depth prediction network \cite{Ranftl2020} pre-trained on Omnidata \cite{eftekhar2021omnidata}. To predict a normalized depth prediction confidence score per pixel $w_{ij}$, we add another network on top of it (refer supplementary for details), which takes both images and depth as input.
\begin{equation}
    \sum_{i}\sum_{j}w_{ij}\cdot \mathcal{L}(d_{ij},\hat{d}_{ij}); ~\textrm{where}, w_{ij} = \phi_w(\mathcal{I}_s,\mathcal{D}_s)_{ij}.
\end{equation}
$\mathcal{L}$ symbolizes the $l_2$ loss between the known pixel depth $d_{ij}$ and the predicted depth $\hat{d}_{ij}$ for $(i, j)$ pixel in the 2D depth image. We call network $\phi_w$ as the ``confidence-prediction head''. 
We take the depths predicted by this network and fuse them with the source image using the Channel Exchange Network \cite{wang2020deep}. The fused information is then projected into $N$ dimensional feature space using network $\mathcal{F}_\theta$, which is a CNN-based architecture. To compute the feature corresponding to a pixel \textbf{p} in the target view \textbf{u}, we warp the features predicted for each source views $\text{\textbf{v}}_k$ to this target view but now using the monocular depth $d \in \mathcal{D}_s $ of the source view.

\begin{equation}
    g_k(\textbf{\text{u}}, \textbf{\text{p}}) = f^{m}_{k}(\mathcal{W}(\textbf{\text{p}}, \textbf{\text{u}}, \textbf{\text{v}}_k, d))
\end{equation}
where, $g_k$ denotes the feature warping function, $f_{k}^{m}(p)$ corresponds to feature in the tensor $\mathcal{F}_\theta(\mathcal{I}_{s}^k, d)$ (corresponding to $k^{th}$ source image and corresponding monocular depth) at pixel $\textbf{p}$, and $\mathcal{W}$ is the warping function.
We now aggregate the warped features corresponding to each source image using a weighted sum based on confidence scores:
\begin{equation}
    \phi_a(\textbf{\text{u}}, \{(\textbf{\text{v}}_k, f_{k}^m(\textbf{\text{p}}))\}) = \sum_{k=1}^K c_k(\textbf{p}) f_{k}^m(\textbf{p})
\end{equation}
here, $c_k$ symbolizes the predicted depth confidence map in the $k^{th}$ view.

\smallskip
\formattedparagraph{\textit{(iii)} Structure Feature Aggregation Per Scene Point.} Given we have features from two different modalities \ie, monocular depth and MVS, we now propose an aggregated feature representation for the target view per scene point $h_{f} = h_\theta(h_m, h_s)$, where $h_\theta$ is a CNN-based neural network with parameters $\theta$ and $h_m$ is the final representation for the monocular estimation and $h_s$ for the stereo-estimation. We aim to attain maximal correlation from both of the input representations \ie, 
\begin{equation}\label{eq:corr}
    \lambda_1\mathcal{C}(h_{f}, h_m) \cdot  \mathcal{C}(h_{f}, h_s) + \lambda_2\mathcal{C}(h_{f}, h_s)+  \lambda_3\mathcal{C}(h_{f}, h_m)
\end{equation}
%
where, $\sum_{i=1}^3 \lambda_i = 1; \lambda_i > 0 ~\forall ~i \in \{1, 2, 3\}$ and $\mathcal{C}(.)$ being the standard correlation function \cite{chandar2016correlational}.
We set $\lambda_1=1$ for nearby points where confidence of monocular depth could be greater than a defined threshold ($\tau$). $\lambda_2 = 1$ for nearby points where confidence of monocular depth is lower than a defined threshold and finally $\lambda_3 = 1$ for far points whose relative depth is greater than a pre-defined threshold ($\sigma$). 
Such a choice is made since stereo features might be less accurate in certain depth range due to low disparity or too close to the lens. The final aggregation $h_f$ comprises of CNN networks $\phi_\alpha$ and $\phi_\beta$, parameterized by $\alpha$ and $\beta$, respectively. The transformed features from each of these modalities are then fed to a CNN network $\phi_\nu$ with parameters $\nu$. Overall aggregated features per scene point is represented as
\begin{equation}\label{eq:asf}
   h_{f} = h_\theta(h_m, h_s) = \phi_\nu(\phi_\alpha(h_m), \phi_\beta(h_s)).
\end{equation}

\begin{figure*}[t]
    \centering
    \includegraphics[width=0.96\linewidth]{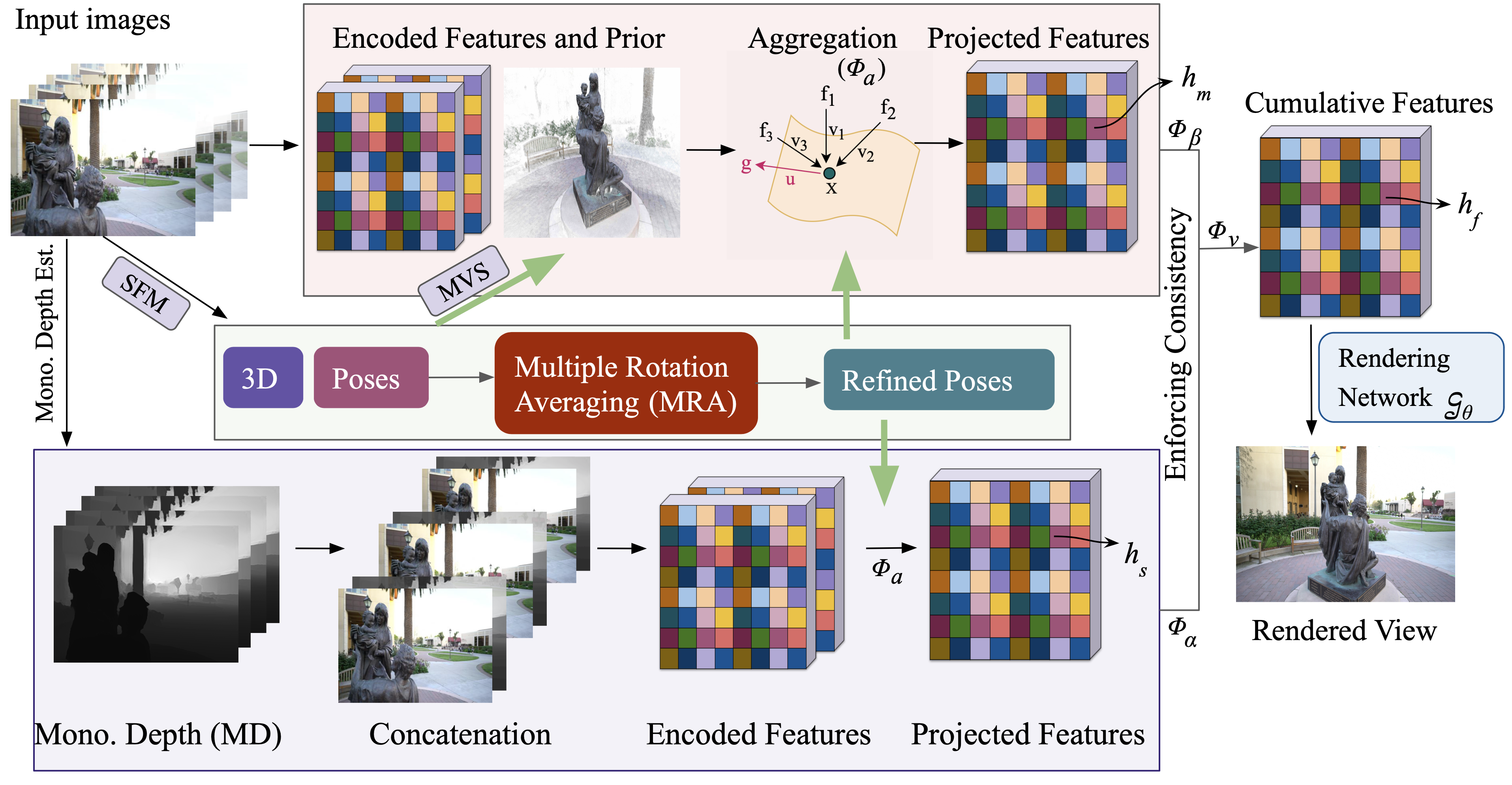}
    \caption{\textbf{Overview.} For a given set of input images, we first estimate the scene 3D structure and the initial value of camera poses using structure from motion (SFM) followed by MVS. We also estimate per-image depth using an existing Monocular Depth estimation model \cite{eftekhar2021omnidata}. Next,
    we generate two sets of Projected Features for each target view: The first set (upper stream) by encoding the input images and then unprojecting the features onto the scaffold, followed by re-projection to the target view and aggregation. For the second set (lower stream), the images are first concatenated with their monocular depths (MD), and the encoded features are then projected using these depths, followed by aggregation. These two feature sets are then merged into Cumulative Features, and finally, the target view is rendered. 
    }\label{fig:pipeline}
\end{figure*}

\subsubsection{Camera Pose Estimation} 
Given we have initially estimated pose-set $\mathcal{P}$, we use multiple motion averaging (MRA) to recover better camera poses. MRA is fast, efficient, and robust in camera pose estimation from noisy key-point correspondences and works well even for sparse image sets. MRA takes a global view of the camera recovery problem. A view-graph corresponding to the initial set of camera poses is constructed as a global representation, with each graph node as a pose. For pose optimization, MRA first performs rotation averaging to recover rotation and then solves for translation. Assume a directed view-graph $\mathcal{G} = (\mathcal{V, E})$. A vertex $\mathcal{V}_j \in \mathcal{V}$ in this view graph corresponds to $j^{th}$ camera absolute rotation ${R}_{j}$ and $\mathcal{E}_{ij} \in \mathcal{E}$ corresponds to the relative orientation $\Tilde{R}_{ij}$ between view $i ~\text{and} ~j$. For our problem, relative orientations are given (noisy), and we aim to solve for absolute pose ${R}_{j}$ robustly. Conventionally, in the presence of noise, the solution is obtained by solving the following optimization problem to satisfy compatibility criteria.
\begin{equation}
\label{eq:rotation_avg_classic}
     \underset{\{{R}_{j}\} }{\text{argmin}} \sum_{\mathcal{E}_{ij} \in \mathcal{E}} \rho\Big( \gamma(\Tilde{R}_{ij}, {R}_j {R}_i^{-1}) \Big)
\end{equation}
where, $\gamma(.)$ denotes a metric on $SO(3)$ to measure distance between two rotation and $\rho(.)$ is a robust $l_1$ norm. We use a graph-neural network to predict the robustified pose view-graph of the scene following the camera pose refining approach of NeuRoRA \cite{purkait2020neurora}. First, the camera poses are checked for outliers via a cyclic consistency in the pose-graph \cite{hartley2013rotation}. Then, a view graph is initialized based on these filtered poses. This view graph is then optimized using the camera pose-refining network of NeuRoRA \cite{purkait2020neurora}.

\subsection{Rendering novel views}
The aggregated feature tensor $\mathcal{F}^a$ along a direction $\textbf{u}$ comprises of the features $h_f$ (Eq. \ref{eq:asf}) for each pixel along $\textbf{u}$, obtained using 3D scene features (discussed in Sec. \ref{sec:scene_feature}). This tensor is now projected to the image space function using $\mathcal{G}_{\theta}$, a CNN-based network with parameters $\theta$. It is quite possible that the regions in the target image may not be encountered in any of the source images. For those points, the values in feature tensor $\mathcal{F}$ are set to $\mathbf{0}$, and thus, they require some inpainting. Denoting the set of test images by $\mathcal{I}_t$, the rendered image corresponding to the $k^{th}$ image in the test set, $\hat{\mathcal{I}}_{t}^{k}$, is predicted based on the following equation:

\begin{equation}
    \hat{\mathcal{I}}_{t}^{k} = \mathcal{G}_\theta(\mathcal{F}^a_k)
    \label{eq:render}
\end{equation}
Similar to stable view synthesis, we parameterize the $\mathcal{G}_\theta$ function using a U-Net \cite{ronneberger2015u} style model which also deals with inpainting/hallucinating the newly discovered regions in the test image. Fig.(\ref{fig:pipeline}) shows the complete overview of the proposed method.

 \begin{table*}[t]
\centering
\scriptsize
\resizebox{\textwidth}{!}
{
    \begin{tabular}{ccccccccccccc}
        \hline
        \multicolumn{1}{c|}{}                                        & \multicolumn{3}{c|}{Truck}                         &  \multicolumn{3}{c|}{M60}     & \multicolumn{3}{c|}{Playground}   & \multicolumn{3}{c}{Train}                      \\ 
        \multicolumn{1}{c|}{}                                                                          & \multicolumn{1}{c}{{\fontsize{6.5}{4}\selectfont PSNR$\uparrow$}} & \multicolumn{1}{c}{{\fontsize{6.5}{4}\selectfont LPIPS$\downarrow$}} & \multicolumn{1}{c|}{{\fontsize{6.5}{4}\selectfont SSIM$\uparrow$}}                                                                                                                                           & \multicolumn{1}{c}{{\fontsize{6.5}{4}\selectfont PSNR$\uparrow$}} & \multicolumn{1}{c}{{\fontsize{6.5}{4}\selectfont LPIPS$\downarrow$}} & \multicolumn{1}{c|}{{\fontsize{6.5}{4}\selectfont SSIM$\uparrow$}}                           & \multicolumn{1}{c}{{\fontsize{6.5}{4}\selectfont PSNR$\uparrow$}} & \multicolumn{1}{c}{{\fontsize{6.5}{4}\selectfont LPIPS$\downarrow$}} & \multicolumn{1}{c|}{{\fontsize{6.5}{4}\selectfont SSIM$\uparrow$}}    & \multicolumn{1}{c}{{\fontsize{6.5}{4}\selectfont PSNR$\uparrow$}} & \multicolumn{1}{c}{{\fontsize{6.5}{4}\selectfont LPIPS$\downarrow$}} & \multicolumn{1}{c}{{\fontsize{6.5}{4}\selectfont SSIM$\uparrow$}}    
        
        \\ \hline
        
        \multicolumn{1}{c|}{\begin{tabular}[c]{@{}c@{}}NeRF++\cite{zhang2020nerf++}\end{tabular}} & 21.8 & 0.31   & \multicolumn{1}{c|}{0.81} &  17.6 &  0.43 & \multicolumn{1}{c|}{0.73} & 21.9  & 0.39 & \multicolumn{1}{c|}{0.79}    &  17.6  & 0.48 & \multicolumn{1}{c}{ 0.68}    \\ 
        \multicolumn{1}{c|}{\begin{tabular}[c|]{@{}c@{}}FVS\cite{riegler2020free}\end{tabular}}          & 21.9  & 0.14 & \multicolumn{1}{c|}{0.84} & 15.8 & 0.32 & \multicolumn{1}{c|}{0.77}  & 21.7  & 0.21 & \multicolumn{1}{c|}{0.83}  & 17.3  & 0.28 & \multicolumn{1}{c}{0.75}                   \\     
        
        \multicolumn{1}{c|}{\begin{tabular}[c]{@{}c@{}}SC-NeRF\cite{jeong2021self}\end{tabular}} & 22.3 & 0.29   & \multicolumn{1}{c|}{0.82} &  18.4 &  0.40 & \multicolumn{1}{c|}{0.76} & 22.4  & 0.35 & \multicolumn{1}{c|}{0.83}    &  18.2  & 0.42 & \multicolumn{1}{c}{ 0.73}    \\ 
        \multicolumn{1}{c|}{\begin{tabular}[c]{@{}c@{}}Point-NeRF\cite{xu2022point} \end{tabular}} & 22.7 & 0.14   & \multicolumn{1}{c|}{0.87} &  19.6 &  0.21 & \multicolumn{1}{c|}{0.85} & 22.2  & 0.19 & \multicolumn{1}{c|}{0.83}    &  18.6  & 0.16 & \multicolumn{1}{c}{ 0.83}    \\ 
         \multicolumn{1}{c|}{\begin{tabular}[c]{@{}c@{}}SVS \cite{riegler2021stable}\end{tabular}} & 22.9 & 0.12   & \multicolumn{1}{c|}{0.88} &  19.1 &  0.22 & \multicolumn{1}{c|}{0.83} & 22.9  & 0.17 & \multicolumn{1}{c|}{0.86}    &  17.9  & 0.19 & \multicolumn{1}{c}{ 0.81}    \\ 
         \hline
        \multicolumn{1}{c|}{\begin{tabular}[c]{@{}c@{}}Ours\end{tabular}} &  \textbf{24.1} &  \textbf{0.12}   & \multicolumn{1}{c|}{ \textbf{0.90}} &  \textbf{20.8} &   \textbf{0.20} & \multicolumn{1}{c|}{ \textbf{0.89}} &  \textbf{23.9}  &  \textbf{0.14} & \multicolumn{1}{c|}{ \textbf{0.90}} &  \textbf{20.1}  &  \textbf{0.13} & \multicolumn{1}{c}{ \textbf{0.88}} \\ %
      
    \end{tabular}
}
\caption{\small Rendered image quality comparison with current state-of-the-art methods in novel view synthesis on the popular Tanks and Temples dataset \cite{knapitsch2017tanks}. We use the popular metrics i.e., PSNR, LPIPS and SSIM for the comparison.}\label{table:tanks}
\end{table*}

\begin{table*}[t]
\centering
\scriptsize
\resizebox{\textwidth}{!}
{
    \begin{tabular}{ccccccccccccc}
        \hline
        \multicolumn{1}{c|}{}                                        & \multicolumn{2}{c|}{Bike}                         &  \multicolumn{2}{c|}{Flowers}     & \multicolumn{2}{c|}{Pirate}   & \multicolumn{2}{c|}{Digger}    &    \multicolumn{2}{c|}{Sandbox} & \multicolumn{2}{c}{Soccertable}               \\ 
        \multicolumn{1}{c|}{}            &                                                     \multicolumn{1}{c}{{\fontsize{6.5}{4}\selectfont SSIM$\uparrow$}} & \multicolumn{1}{c|}{{\fontsize{6.5}{4}\selectfont LPIPS$\downarrow$}}                                                                                                                                            & \multicolumn{1}{c}{{\fontsize{6.5}{4}\selectfont SSIM$\uparrow$}} & \multicolumn{1}{c|}{{\fontsize{6.5}{4}\selectfont LPIPS$\downarrow$}}                           & \multicolumn{1}{c}{{\fontsize{6.5}{4}\selectfont SSIM$\uparrow$}} & \multicolumn{1}{c|}{{\fontsize{6.5}{4}\selectfont LPIPS$\downarrow$}}    & \multicolumn{1}{c}{{\fontsize{6.5}{4}\selectfont SSIM$\uparrow$}} & \multicolumn{1}{c|}{{\fontsize{6.5}{4}\selectfont LPIPS$\downarrow$}}    
        & \multicolumn{1}{c}{{\fontsize{6.5}{4}\selectfont SSIM$\uparrow$}} & \multicolumn{1}{c|}{{\fontsize{6.5}{4}\selectfont LPIPS$\downarrow$}}  
        & \multicolumn{1}{c}{{\fontsize{6.5}{4}\selectfont SSIM$\uparrow$}} & \multicolumn{1}{c}{{\fontsize{6.5}{4}\selectfont LPIPS$\downarrow$}}  
        
        \\ \hline
        
        \multicolumn{1}{c|}{\begin{tabular}[c]{@{}c@{}}NeRF++\cite{zhang2020nerf++}\end{tabular}}  & 0.71   & \multicolumn{1}{c|}{0.27}  &  0.80 & \multicolumn{1}{c|}{0.31} &  0.71 & \multicolumn{1}{c|}{0.43}      & 0.65 & \multicolumn{1}{c|}{ 0.35} & 0.84 & \multicolumn{1}{c|}{ 0.24} & 0.87 & \multicolumn{1}{c}{ 0.21}    \\ 
        \multicolumn{1}{c|}{\begin{tabular}[c]{@{}c@{}}FVS\cite{riegler2020free}\end{tabular}}  & 0.61   & \multicolumn{1}{c|}{0.28}  &  0.79 & \multicolumn{1}{c|}{0.27} &  0.69 & \multicolumn{1}{c|}{0.37}      & 0.68 & \multicolumn{1}{c|}{ 0.24} & 0.78 & \multicolumn{1}{c|}{ 0.32} & 0.82 & \multicolumn{1}{c}{ 0.21}                  \\

         \multicolumn{1}{c|}{\begin{tabular}[c]{@{}c@{}}SVS\cite{riegler2021stable}\end{tabular}}  & 0.74   & \multicolumn{1}{c|}{0.22}  &  0.84 & \multicolumn{1}{c|}{0.21} &  0.75 & \multicolumn{1}{c|}{0.32}      & 0.77 & \multicolumn{1}{c|}{ 0.18} & 0.85 & \multicolumn{1}{c|}{ 0.20} & 0.91 & \multicolumn{1}{c}{ 0.15}     \\ 
         \hline
        \multicolumn{1}{c|}{\begin{tabular}[c]{@{}c@{}}Ours\end{tabular}}  & \textbf{0.79}   & \multicolumn{1}{c|}{\textbf{0.19}}  &  \textbf{0.88} & \multicolumn{1}{c|}{\textbf{0.18}} &  \textbf{0.80} & \multicolumn{1}{c|}{\textbf{0.29}}      & \textbf{0.81} & \multicolumn{1}{c|}{ \textbf{0.16}} & \textbf{0.91} & \multicolumn{1}{c|}{ \textbf{0.17}} & \textbf{0.94} & \multicolumn{1}{c}{ \textbf{0.13}}  \\ %
      
    \end{tabular}
}
\caption{\small Quantitative comparison on FVS dataset \cite{riegler2020free}. We use the popular metrics, \ie, PSNR, LPIPS and SSIM for the comparison.}
\label{table:fvs}
\end{table*}

\begin{figure*}[t]
    \centering
    \includegraphics[scale=0.30]{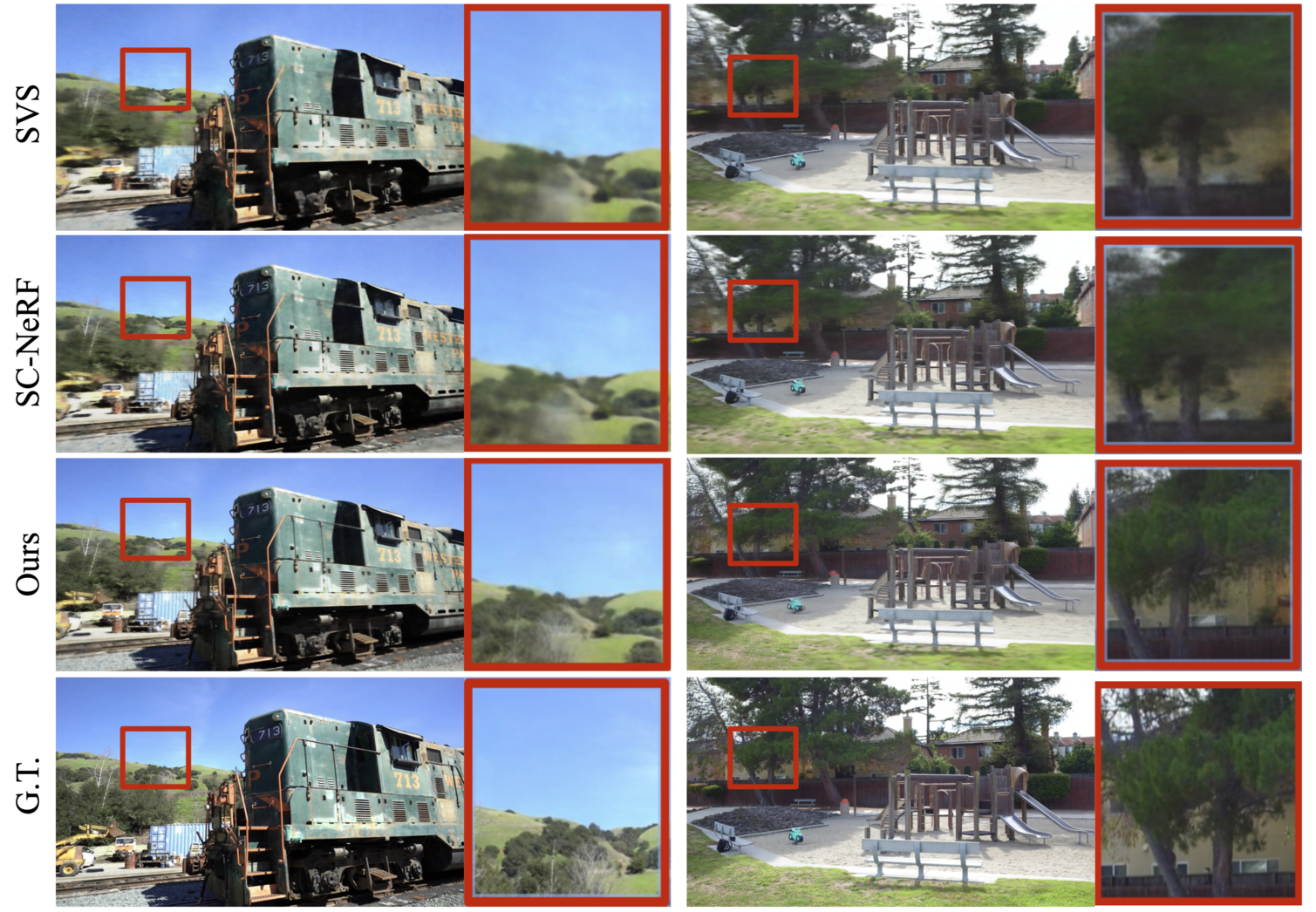}
    \caption{Qualitative comparison on tanks and temples dataset. If we zoom into the scene details, our approach results show considerably less artifacts than the state-of-the-art methods enabling unparalleled level of realism in image-based rendering. Our PSNR values for the above two scenes are (\textbf{20.1}, \textbf{23.9}). In comparison, SC-NeRF\cite{jeong2021self} and SVS \cite{riegler2021stable} provide (\textbf{17.3}, \textbf{22.4}) and (\textbf{17.9}, \textbf{22.9}), respectively.}\label{fig:visual_results}
\end{figure*}

\subsection{Joint Optimization}\label{ssec:joint_opt}

\noindent
Given $\mathcal{I}_s$, we train the model using our overall loss function $\mathcal{L}$ comprising of two terms.

\begin{equation}
\mathcal{L} = \mathcal{L}_{s} + \mathcal{L}_{p}  
\label{eq:pose_opt}
\end{equation}

\smallskip
\noindent
\textbf{\textit{(i)}}  The \textbf{first loss term} $\mathcal{L}_s$ encourages the network to learn better features corresponding to the scene point. The $\mathcal{L}_s$ comprises to two objective function $\mathcal{L}_{rgb}$ and $\mathcal{L}_{corr}$. The function $\mathcal{L}_{rgb}$ measures the discrepancies between the rendering of an image in the source set and the actual image available in the set. This objective is used to train the structure estimation and rendering network parameters jointly. Whereas $\mathcal{L}_{corr}$ is used for maximizing the correlation objective discussed in Eq.\eqref{eq:corr} to arrive at the optimal aggregated scene representation.

\begin{equation}
\label{eq:loss_combined}
\begin{aligned}
    \mathcal{L}_s & = \mathcal{L}_{rgb} + \mathcal{L}_{corr}
\end{aligned}
\end{equation}
$\mathcal{L}_{corr}$ takes the negative of the objective defined in Eq.\eqref{eq:corr}. The structure network parameters corresponding to functions $f_k$ and $f_k^m$ (cf. Sec. \ref{sec:scene_feature}) and is updated using $\mathcal{L}_{corr}$ and $\mathcal{L}_{rgb}$. The rendering network parameters corresponding to $\mathcal{G}_\theta$ updated using only $\mathcal{L}_{rgb}$.

\smallskip
\noindent
\textbf{\textit{(ii)}} The \textbf{second loss term} $\mathcal{L}_p$ corresponds to the camera pose refinement. We used the following loss $\mathcal{L}_p$ to improve the camera pose estimation.
\begin{equation}
\mathcal{L}_p = \mathcal{L}_{mra} 
\label{eq:pose_opt}
\end{equation}
As is known, the multiple motion averaging and the color rendering cost functions are directly impacted by the camera pose parameters. And therefore, the $\mathcal{L}_{rgb}$ is inherently used as an intermediate objective in which the camera pose term is constant between the two different representations.

\section{Experiments, Results and Ablations}
Here, we discuss our experimental results and their comparison to relevant baselines. Later, we provide critical ablation analysis, a study of our camera pose estimation idea, and 3D structure recovery. Finally, the section concludes with a discussion of a few extensions of our approach\footnote{Please refer supplementary for the train-test setup, implementation details regarding hyperparameters, and architectures to reproduce our results.}.

\smallskip
\formattedparagraph{Baselines.} We compare our approach with the recently proposed 
novel view synthesis methods that work well in practice for real-world scenes. Our baseline list includes NeRF++\cite{zhang2020nerf++}, SC-NeRF \cite{jeong2021self}, PointNeRF \cite{xu2022point}, FVS\cite{riegler2020free} and SVS \cite{riegler2021stable}. All baseline methods are trained on the source image set of the testing sequence. Our method's train and test set are the same as SVS \cite{riegler2021stable}.

\subsection{Dataset and Results}

For evaluation, we used popular benchmark datasets comprising real-world and synthetic datasets. Namely, we used the Tanks and Temples \cite{knapitsch2017tanks}, FVS \cite{riegler2020free}, Mip-NeRF 360  \cite{barron2022mip}, and the DTU \cite{jensen2014large} dataset. The first two datasets contain real-world scenes, whereas the last is an object-centric synthetic benchmark dataset. Although our approach mainly targets realistic scenes, for completeness, we performed and tabulated results on a synthetic object-based DTU dataset (see supplementary). Next, we discuss datasets and results.

\smallskip
\formattedparagraph{\textit{(i)} Tanks and Temples Dataset.}
It consists of images of large-scale real-world scenes taken from a freely moving camera, consisting of indoor and outdoor scenes. Unfortunately, we have no ground-truth poses; thus, estimated poses using COLMAP \cite{schoenberger2016sfm} are treated as pseudo-ground-truth poses. The dataset consists of a total of 21 scenes. Similar to the SVS setup, we use 15 of 21 scenes for training, 2 for validation, and the rest as test scenes. We also use a disjoint set of target views from the test scenes to evaluate all the methods. The source views in test scenes are used for scene-specific fine-tuning if required. Table \ref{table:tanks} shows the statistical comparison of our approach with competing baselines. Our approach consistently outperforms the relevant baselines on this challenging dataset. Furthermore, if we zoom into the scene's finer details, our approach clearly preserves details better than other approaches (see Fig. \ref{fig:visual_results}).

\smallskip
\formattedparagraph{\textit{(ii)} FVS Dataset.} It comprises six real-world scenes. Each scene was recorded two or more times to completely differentiate the source image set to form the target image set. Table \ref{table:fvs} summarises this dataset's view synthesis quantitative results. We compare our method with the state-of-the-art NeRF++ \cite{zhang2020nerf++}, Free-View synthesis \cite{riegler2020free}, and SVS \cite{riegler2021stable} methods. We observe that FVS improves over NeRF++, whereas SVS results on this dataset do not consistently outperform other previous methods. On the contrary, our method consistently provides better results than prior art in all categories. The dataset's sample images and its qualitative results are provided in the supplementary.

\smallskip
\formattedparagraph{\textit{(iii)} Mip-NeRF 360 Dataset.} We further test our approach on this recently proposed dataset comprising unbounded scenes \cite{barron2022mip}. We use the train-test setup per scene proposed by the authors to fine-tune our network and compare it with Mip-NeRF 360 \cite{barron2022mip}, NeRF++ \cite{zhang2020nerf++}, and SVS \cite{riegler2021stable} baseline. Table \ref{table:mipnerf} shows the results for this comparison. Clearly, our methodical take on this problem generalizes well, outperforming other baseline methods for most examples.

\begin{table}[h]
\centering
\footnotesize   
\begin{tabular}{c|ccc } 
\hline
 & PSNR$\uparrow$ & LPIPS$\downarrow$ & SSIM$\uparrow$ \\
 \hline
 NeRF++\cite{zhang2020nerf++} & 25.03 & 0.355 & 0.682 \\ 
 SVS\cite{riegler2021stable} & 25.28 & 0.218 & 0.783\\
 \hline
 PBNR\cite{kopanas2021point} & 23.55 & 0.262 & 0.722 \\ 
 Mip-NeRF 360\cite{barron2022mip} & 27.07 & 0.251 & 0.781 \\ 
 \hline
 Ours & \textbf{27.95} & \textbf{0.232} & \textbf{0.797} \\ 
\end{tabular}
\caption{\small Quantitative comparison of our approach with existing baselines on the Mip-NeRF 360 dataset.}
\label{table:mipnerf}
\end{table}

\subsection{Ablation Analysis}

\begin{figure}[t]
    \centering
    \includegraphics[width=0.49\linewidth]{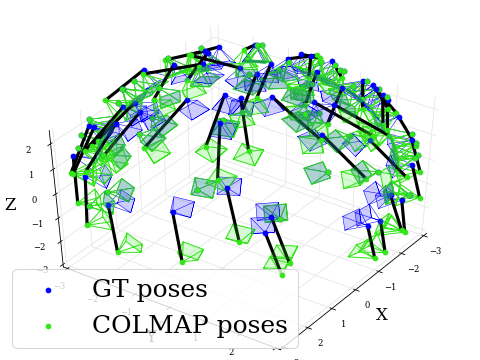}
    \includegraphics[width=0.49\linewidth]{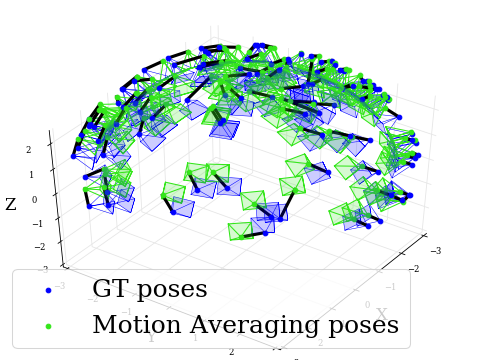}
    \caption{\footnotesize COLMAP \cite{schoenberger2016sfm} and our method's camera poses on Lego dataset \cite{mildenhall2020nerf}. Here, we add some noise to the pairwise matched correspondences to simulate realistic scenarios. \textbf{Left:} Estimated camera pose from COLMAP and the ground truth pose. Here, black line shows the pose error. \textbf{Right:} Our recovered camera pose. It is easy to infer our approach robustness.}\label{fig:correspondences}
\end{figure}

\begin{figure}[h]
    \centering
    \includegraphics[scale=0.5]{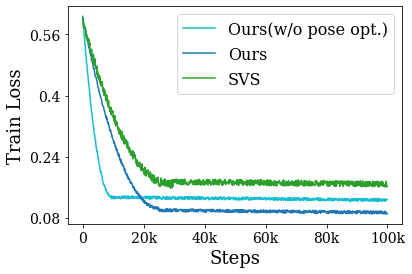}
    \caption{ \footnotesize Training loss curve of our method with and without pose optimization compared to state-of-the-art SVS \cite{riegler2021stable}}\label{fig:opt}
\end{figure}

\smallskip
\formattedparagraph{\textit{(i)} Effect of our camera-pose estimation approach.} We performed this experiment to show the benefit of using multiple motion averaging (MRA) instead of relying only on COLMAP poses and taking them for granted. For this experiment, we used the popular Lego dataset \cite{mildenhall2020nerf}. We first compute the COLMAP camera poses shown in Fig.\ref{fig:correspondences} (\textbf{Left}). Next, we use this initial camera pose set to perform MRA utilizing the view-graph optimization. Our recovered camera poses are shown in Fig.\ref{fig:correspondences} (\textbf{Right}). For clarity, we also show the ground-truth camera pose frustum and the error between the ground-truth pose and recovered pose for the two respective cases. It is clear from Fig.(\ref{fig:correspondences}) that a principled approach is required for better camera-pose recovery in view-synthesis problem.
Furthermore, to show the benefit of our estimated camera pose in novel-view synthesis, we show our network training loss curve with and without our camera pose estimation approach in Fig.(\ref{fig:opt}). For clarity, we also plot the SVS training curve. Firstly, our method shows better loss response due to the improved 3D geometric scaffold of the scene. Moreover, we improve on training loss further by introducing neural graph-based MRA, showing the benefit of our proposed loss function.

\begin{figure}
    \centering
    \includegraphics[scale=0.19]{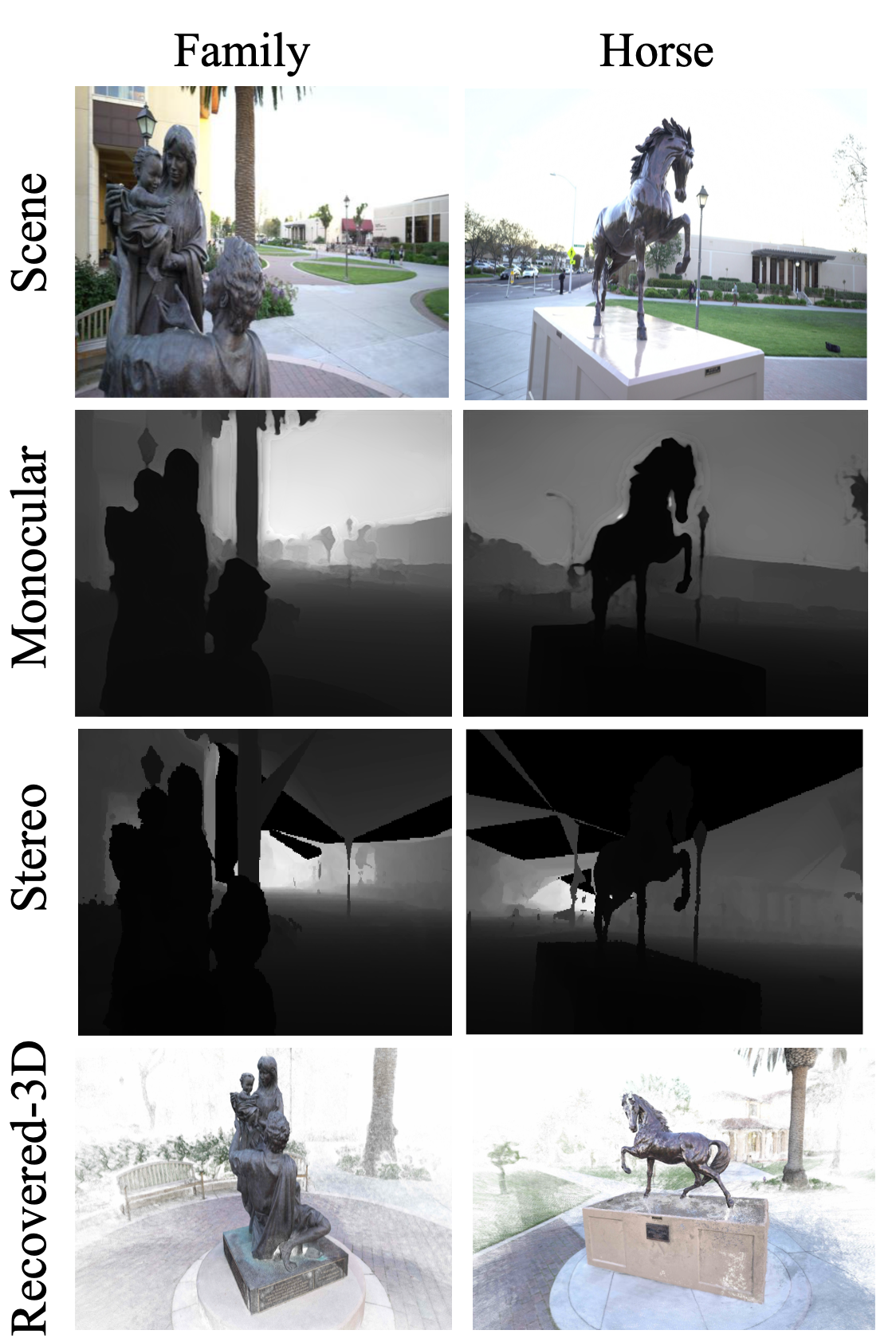}
    \caption{Comparison of depth predicted by monocular and stereo based method on the scenes from Tanks and Temples dataset.}
    \label{fig:depth}
\end{figure}

\smallskip
\formattedparagraph{\textit{(ii)} Effect of using MVS with monocular depth.} 
We performed this ablation to demonstrate the effectiveness of our approach in recovering geometric reconstruction of the scene. Fig.(\ref{fig:depth}) clearly shows the complementary nature of the monocular depth and stereo depth in 3D reconstruction from images. While MVS provides reliable depth results for near and mid-range points, monocular depth can deliver a good depth inference (up to scale) on the far points in the scene. By carefully integrating both modalities' depth, we have better 3D information about the scene. Fig.(\ref{fig:depth}) shows our results on a couple of scenes from Tanks and Temples dataset, demonstrating the suitability of our approach.

\section{Conclusion}\label{sec:conclusion}
The approach presented in this paper directly addresses the shortcomings of the currently popular methods in novel view synthesis. Our approach integrates concepts from the learning-based approaches and classical techniques in multiple-view geometry for solving novel-view synthesis. We demonstrate that by 
exploiting the \textbf{\textit{(i)}} complementary nature of monocular depth estimation and multiple view stereo (MVS) in the 3D reconstruction of the scene, and \textbf{\textit{(ii)}} usefulness of multiple rotation averaging in structure from motion, we can achieve better-rendering quality than other dominant approaches in novel view synthesis. We confirm the effectiveness of our approach via a comprehensive evaluation and quantitative comparison with baselines on several benchmark datasets. Although our work enabled improved photorealistic rendering, several exciting avenues exist for further improvement. One interesting future extension is fully automating our introduced view synthesis pipeline, \ie, adding a learnable MVS framework to estimate intrinsic, extrinsic camera parameters with scene reconstruction for better novel view synthesis.

{\small
\bibliographystyle{ieee_fullname}
\bibliography{egbib}
}


\twocolumn[\section*{\centering \Large Enhanced Stable View Synthesis \\ ---Supplementary Material---\\}]

\input{supplementary.tex}

\end{document}

%% file: supplementary.tex
\begin{abstract}
    This draft accompanies the main paper. It provides more experimental results showing the suitability of our proposed approach. Furthermore, it discusses the graph neural network-based multiple rotation averaging and our software implementation details.
\end{abstract}

\section{Synthetic Objects}
We further evaluate our proposed method on a synthetic object-centric dataset to investigate its advantages in cases where all points lie within a threshold distance from the camera. The aim is to examine whether integrating a monocular depth with the estimated stereo structure is helpful for these cases. We use the same setup as SVS\cite{riegler2021stable} for evaluation, separating ten images as target novel images and use the remaining 39 as the source images for evaluating interpolation and extrapolation. Table \ref{table:dtu} compares all the baselines and our method on this dataset. It consists of results for both view interpolation (left value) and extrapolation setups (right value) for this dataset as done in the SVS paper\cite{riegler2021stable}.
It can be observed that even in this case, which doesn't involve \textit{far-away} points, our method is either similar in performance or performs better, especially in terms of PSNR values. Also, performance gap on the extrapolation task is marginally higher than the interpolation counterpart, for the PSNR values. This further highlights the importance of our method for the nearby region where we try to maximize the consistency between RGB-D features and stero-estimated projection of image features, for places where the monocular network is highly confident.

\begin{table*}[t]
\centering
\scriptsize
\resizebox{\textwidth}{!}
{
    \begin{tabular}{cccccccccc}
        \hline
        \multicolumn{1}{c|}{}                                        & \multicolumn{3}{c|}{65}                         &  \multicolumn{3}{c|}{106}     & \multicolumn{3}{c}{118}                         \\ 
        \multicolumn{1}{c|}{}                                                                          & \multicolumn{1}{c}{{\fontsize{6.5}{4}\selectfont PSNR$\uparrow$}} & \multicolumn{1}{c}{{\fontsize{6.5}{4}\selectfont LPIPS$\downarrow$}} & \multicolumn{1}{c|}{{\fontsize{6.5}{4}\selectfont SSIM$\uparrow$}}                                                                                                                                           & \multicolumn{1}{c}{{\fontsize{6.5}{4}\selectfont PSNR$\uparrow$}} & \multicolumn{1}{c}{{\fontsize{6.5}{4}\selectfont LPIPS$\downarrow$}} & \multicolumn{1}{c|}{{\fontsize{6.5}{4}\selectfont SSIM$\uparrow$}}                           & \multicolumn{1}{c}{{\fontsize{6.5}{4}\selectfont PSNR$\uparrow$}} & \multicolumn{1}{c}{{\fontsize{6.5}{4}\selectfont LPIPS$\downarrow$}} & \multicolumn{1}{c}{{\fontsize{6.5}{4}\selectfont SSIM$\uparrow$}}    
        
        \\ \hline
        
        \multicolumn{1}{c|}{\begin{tabular}[c|]{@{}c@{}}FVS\cite{riegler2020free}\end{tabular}}          & 30.1/25.2  & 0.03/0.07 & \multicolumn{1}{c|}{0.97/0.95} & 32.3/27.1 & 0.03/0.08 & \multicolumn{1}{c|}{0.95/0.93}  & 34.9/28.7  & 0.02/0.07 & \multicolumn{1}{c}{0.97/0.94}                    \\     
        
         \multicolumn{1}{c|}{\begin{tabular}[c]{@{}c@{}}SVS \cite{riegler2021stable}\end{tabular}} & 31.9/26.1 & {0.02/0.06}   & \multicolumn{1}{c|}{{0.97}/0.95} &  33.8/29.7 &  0.02/0.04 & \multicolumn{1}{c|}{0.98/0.95} & 36.7/30.8  & {0.02/0.05} & \multicolumn{1}{c}{{0.97/0.96}}        \\ 
         \hline
        \multicolumn{1}{c|}{\begin{tabular}[c]{@{}c@{}}Ours\end{tabular}} &  {32.4/26.9} &  {0.03/0.07}   & \multicolumn{1}{c|}{ {0.98/0.96}} &  {34.3/30.6} &   {0.02/0.03} & \multicolumn{1}{c|}{ {0.98/0.96}} &  {37.1/31.3}  &  {0.02/0.05} & \multicolumn{1}{c}{ {0.97/0.96}}    \\ %
      
    \end{tabular}
}
\caption{\small Performance comparison with on DTU dataset\cite{hedman2016scalable}. We use the popular metrics i.e., PSNR, LPIPS and SSIM for the comparison.}\label{table:dtu}
\end{table*}

\begin{table*}[t]
\centering
\scriptsize
\resizebox{\textwidth}{!}
{
    \begin{tabular}{ccccccccccccc}
        \hline
        \multicolumn{1}{c|}{}                                        & \multicolumn{3}{c|}{Truck}                         &  \multicolumn{3}{c|}{M60}     & \multicolumn{3}{c|}{Playground}   & \multicolumn{3}{c}{Train}                      \\ 
        \multicolumn{1}{c|}{}                                                                          & \multicolumn{1}{c}{{\fontsize{6.5}{4}\selectfont PSNR$\uparrow$}} & \multicolumn{1}{c}{{\fontsize{6.5}{4}\selectfont LPIPS$\downarrow$}} & \multicolumn{1}{c|}{{\fontsize{6.5}{4}\selectfont SSIM$\uparrow$}}                                                                                                                                           & \multicolumn{1}{c}{{\fontsize{6.5}{4}\selectfont PSNR$\uparrow$}} & \multicolumn{1}{c}{{\fontsize{6.5}{4}\selectfont LPIPS$\downarrow$}} & \multicolumn{1}{c|}{{\fontsize{6.5}{4}\selectfont SSIM$\uparrow$}}                           & \multicolumn{1}{c}{{\fontsize{6.5}{4}\selectfont PSNR$\uparrow$}} & \multicolumn{1}{c}{{\fontsize{6.5}{4}\selectfont LPIPS$\downarrow$}} & \multicolumn{1}{c|}{{\fontsize{6.5}{4}\selectfont SSIM$\uparrow$}}    & \multicolumn{1}{c}{{\fontsize{6.5}{4}\selectfont PSNR$\uparrow$}} & \multicolumn{1}{c}{{\fontsize{6.5}{4}\selectfont LPIPS$\downarrow$}} & \multicolumn{1}{c}{{\fontsize{6.5}{4}\selectfont SSIM$\uparrow$}}    
        
        \\ \hline
        
        \multicolumn{1}{c|}{\begin{tabular}[c|]{@{}c@{}}FVS\cite{riegler2020free}\end{tabular}}          & 21.9  & 0.14 & \multicolumn{1}{c|}{0.84} & 15.8 & 0.32 & \multicolumn{1}{c|}{0.77}  & 21.7  & 0.21 & \multicolumn{1}{c|}{0.83}  & 17.3  & 0.28 & \multicolumn{1}{c}{0.75}                   \\     
        
         \multicolumn{1}{c|}{\begin{tabular}[c]{@{}c@{}}SVS \cite{riegler2021stable}\end{tabular}} & 22.2 & 0.16   & \multicolumn{1}{c|}{0.85} &  18.4 &  0.24 & \multicolumn{1}{c|}{0.79} & 22.4  & 0.20 & \multicolumn{1}{c|}{0.83}    &  17.3  & 0.21 & \multicolumn{1}{c}{ 0.79}    \\ 
         \hline
        \multicolumn{1}{c|}{\begin{tabular}[c]{@{}c@{}}Ours\end{tabular}} &  \textbf{23.4} &  \textbf{0.14}   & \multicolumn{1}{c|}{ \textbf{0.88}} &  \textbf{19.6} &   \textbf{0.23} & \multicolumn{1}{c|}{ \textbf{0.85}} &  \textbf{22.6}  &  \textbf{0.18} & \multicolumn{1}{c|}{ \textbf{0.89}} &  \textbf{19.2}  &  \textbf{0.16} & \multicolumn{1}{c}{ \textbf{0.84}} \\ %
      
    \end{tabular}
    }
\caption{\small Performance comparison with {state-of-the-art} methods on Tanks and Temples dataset \cite{knapitsch2017tanks} in a scene-agnostic setup. We use the popular metrics i.e., PSNR, LPIPS and SSIM for the comparison.}\label{table:tanks_agnostic}
\end{table*}

\section{Scene-Agnostic Model}

We analyze the scene-agnostic version of our approach and SVS, also comparing with the FVS\cite{riegler2020free} method. The model is trained on scenes corresponding to training data and then is directly evaluated on a disjoint set of test scenes without any tuning. Table \ref{table:tanks_agnostic} shows the results for this version on the four scenes of the Tanks and Temples data set used in the paper, namely \textit{truck}, \textit{M60}, \textit{playground} and \textit{train}, where the model is trained using the 15 other scenes from this dataset. It can be observed that our approach can offer significantly better results, even in the scene-agnostic setup, when compared with SVS and FVS. Also, for both our method and SVS, the results are improved from scene-specific finetuning compared to the scene-agnostic setup, which can be observed by comparing the statistics presented in Table \ref{table:tanks_agnostic} here and Table \textcolor{red}{1} in the main paper.

\section{Graph Neural Networks for MRA}\label{sec:GNNmra}
We now discuss our pose refining scheme inspired from NeuRoRA \cite{purkait2020neurora}. It uses a Message Passing Neural Network (MPNN) to predict robust poses given a completely initialized view graph. Given the estimated relative rotations using an SFM algorithm, they are used to initialize absolute rotations by fixing a source vertex as the frame of reference and then calculating absolute rotation of each vertex w.r.t. this frame by traversing along the minimum spanning tree. This is followed by a cyclic consistency check to remove outliers. Finally, we have the initialized observed relative rotations and initialized absolute rotations. These comprise a completely initialized view-graph.

\smallskip
\noindent
Now, for each node $k$ in this graph, with neighbouring set denoted by $\mathcal{Q}_k$, the state of this node at step $t$, denoted by $h^t_k$, is generated by processing the aggregated signal feature $s_j^t$ it receives from all the nodes $v \in \mathcal{Q}_k$ and its state at step $t-1$:
\begin{equation}
    h_k^t = \rho(h_k^{t-1},s_k^t) 
\end{equation}
where $\rho$ is some function to process these features jointly. The aggregated signal $s_k^t$ is just a processed combination of updated states of edges corresponding to this node $k$:
\begin{equation}
    s_k^t = \psi^a_{i \in \mathcal{Q}_k} \psi^b(h_k^{t-1}, h_i^{t-1}, r_{ij}) 
\end{equation}
where, $\psi^b$ is a processing function responsible for updating the signal accumulated from the edge between nodes $i$ and $k$, $r_{ij}$ is a feature representation for this edge. This is followed by a differentiable operation $\psi^a$, which can be interpreted as some activation function. For our case, both $\rho$ and $\psi^b$ are concatenation with 1D convolutions and a ReLU activation. Please refer \cite{gilmer2017neural, purkait2020neurora} for further details.

\smallskip
\noindent
%
\textbf{Pose-refining GNN}. 
Given the completely initialized view graph, we denote the rotations corresponding to its vertices as $\Tilde{R}_i$. Also, the edges of these view graph comprise the relative rotations between the 2 nodes. The edge feature $r_{ij}$ described above is the calculated using these observed relative rotation between the two nodes denoted as $\Tilde{R}_{ij}$. The input to the GNN is the set of rotations $\Tilde{R}_i$ and the edge feature $r_{ij}$. This edge feature is calculated as the discrepancy between the initialized absolute rotations $\Tilde{R}_i$ and observed relative rotations $\Tilde{R}_{ij}$ as follows:
\begin{equation}
    r_{ij} = \Tilde{R}^{-1}_{j} \Tilde{R}_{ij} \Tilde{R}_{i}
\end{equation}
This leads to a supervised learning problem for the GNN, where, using the input graph denoted as $\{\Tilde{R}_i,r_{ij}\}$, the aim is to estimate the absolute rotations $\hat{R}_i$ as close as possible to the correct rotations $\{R_i\}$ in the source node frame:
\begin{equation}
    \{\hat{R}_{i}\} = \mathcal{G}(\{\Tilde{R}_{i}\};\Theta)
\end{equation}
where $\Theta$ denote the pose-GNN parameters. The network is trained for this setup using the rotation averaging loss described in the paper. Specifically, the goal is to minimize the discrepancy between observed relative rotations $\Tilde{R}_{ij}$ and estimated relative rotations $\hat{R}_j\hat{R}_{i}^{-1}$. Given only relative rotations are used in this loss function, this it might be same even if any constant angular deviation to the predicted rotations. Thus following NeuRoRA \cite{purkait2020neurora}, we also add a weighted regularizer term to learn a one-to-one mapping between inputs and outputs which minimizes the discrepancy between initialized absolute rotations and predicted absolute rotations. This leads to the following aggregated cost function $\mathcal{L}_{mra}$ for a given graph with $\mathcal{E}$ denoting its edge set and $\mathcal{V}$ denoting the set of nodes:
\begin{equation}
   \mathcal{L}_{mra} =  \sum_{\mathcal{E}_{ij}\in \mathcal{E}}d_{Q}(\hat{R}_j\hat{R}_i^{-1}, \Tilde{R}_{ij}) + \beta \sum_{\mathcal{V}_{i}\in \mathcal{V}}d_Q(\hat{R}_i, \Tilde{R}_i)
\end{equation}
where $d_Q$ is some distance metric between two rotations. We also follow quaternion representation for these rotations similar to NeuRoRA\cite{purkait2020neurora}.

\section{Adapting to other methods}
To further show the effectiveness of our joint feature estimation and pose updating proposition, we experimented with other popular frameworks, namely IbrNet \cite{wang2021ibrnet} and S-IbrNet\cite{sun2022learning}, on two scenes of the Tanks and Temples dataset namely Truck and Playground (P.G.), in our setup. This is done by updating their feature generation modules and integrating our pose-refining module for joint optimization. Table \ref{tab:other_methods} shows the PSNR values for this experiment showing numbers for these methods before and after (E-IbrNet, E-SIbrNet) the integration and also compares these results with the method proposed in the paper. 
It can be observed that both the methods (IbrNet, S-IbrNet) have their PSNR values improved significantly (around 1.7, 0.9 on average across the two scenes) and finally, our proposed method in the paper performs the best on both the scenes.
\begin{table}[!htb]
\footnotesize
    \centering
    \begin{tabular}{c|c|c|c|c|c}
           & IbrNet \cite{wang2021ibrnet} & S-IbrNet \cite{sun2022learning} & E-IbrNet & E-SIbrNet & Ours \\
    \hline
        Truck & 19.7 &22.5 & {21.9} & {23.6} & \textbf{24.1} \\
        P.G. & 22.2 & 23.1 & {23.3} & {23.7}  & \textbf{23.9}\\
    \end{tabular}
    \caption{PSNR comparison. E-IbrBet and E-SIbrNet show the results when our approach is put to \cite{wang2021ibrnet} and \cite{sun2022learning} network design.}
    \label{tab:other_methods}
\end{table}

\section{Ablation on depth threshold ($\sigma$)} 
For all the experiments, we have set the depth threshold value based on our empirical observations. We extensively analyzed this quantity on the Tanks and Temples dataset and have set it as $0.66$ (relative to median depth) for all the datasets used in the paper. Table \ref{tab:depth_threshold} shows the analysis of PSNR values on two scenes of Tanks and Temples dataset (Truck, M60) for various values of this quantity (relative to the median depth).  It can be observed that the best results are at the value of 0.66, thereby providing justification for the selected value.

\begin{table}[!htb]
\footnotesize
    \centering
    \begin{tabular}{c|c|c|c|c|c|c|c}
    
         $\sigma$ & 0.25 & 0.50 & 0.66 & 1.0& 1.25 & 1.5& 2.0\\
        \hline
        Truck & 23.4 &23.8 &\textbf{24.1} &23.6 &23.6 &23.5 &23.5\\
        M60 & 20.4 &20.3 &\textbf{20.8} &20.4 &20.5 &20.2 &20.3 \\
        
    \end{tabular}
    \caption{PSNR values for ablation on the depth threshold value.}
    \label{tab:depth_threshold}
\end{table}

\section{Training and Evaluation details}
We now discuss the implementation details involving the training and evaluation setup along with 
values for parameters involved in our approach. The training is performed on the tanks and temples dataset for our method, SVS\cite{riegler2021stable} and FVS\cite{riegler2020free}.
Then, for the results on all the scenes corresponding to various datasets discussed in the paper, we tune our method and SVS for scene-specific Network fine-tuning as discussed in SVS\cite{riegler2021stable}.
For the baselines including NeRF++ \cite{zhang2020nerf++}, SC-NeRF\cite{jeong2021self} and Point-NeRF\cite{xu2022point} also require per-scene fitting. This scene-specific training/tuning involves using a source image set of that scene for learning and a disjoint test set of images corresponding to the same scene for evaluation. For the scene agnostic scenario, trained models on tanks and temples are directly evaluated on the test set of the scene. Also, the $\mathcal{L}_{rgb}$ loss term, used in the Eq.($\textcolor{red}{13}$) in the main paper, corresponds to the perceptual loss described in Eq.($\textcolor{red}{6}$) in the SVS paper \cite{riegler2021stable}.\\

\noindent
\textbf{Architecture details}. The monocular depth prediction is a DPT\cite{Ranftl2020} architecture trained on Omnidata\cite{eftekhar2021omnidata}. The network for predicting confidence score for each pixel is just-another head starting from fifth last layer of the depth prediction network with the same architecture as that of the depth prediction head. This is trained while keeping the depth prediction network as frozen. Note, the complete architecture is just used for obtaining depth and confidence per pixel and then is linked with rest of the pipeline. Also, the confidence weighted loss involves $l_2$ regularization of the predicted weights to avoid the solutions, where each/most of the weights are assigned to a single/some pixels. 
The network $\mathcal{F}_\theta$ for projecting monocular features consists of a ResNet-50 architecture equipped with the Channel Exchanging Layers \cite{wang2020deep}. 
The functions $\phi_\alpha$, $\phi_\beta$ and $\phi_\mu$ are each 3 layered CNN networks followed by a BatchNorm layer and a ReLU activation.
For the SVS features, the architecture used in the SVS paper is followed involving a U-Net for encoding images.
As discussed in the paper, the rendering network is also a U-Net architecture same as in SVS paper\cite{riegler2021stable}.\\

\noindent
\textbf{Hyperparameter details}. The training is performed using an Adam optimizer setting learning rate to $10^{-3}$, $\beta_1$ to $0.9$, $\beta_2$ to $0.999$ and $\epsilon$ to $10^{-8}$. The model is trained for 600,000 iterations on the training set with batch size of 1 and 3 source images sampled per iteration, following the SVS setup\cite{riegler2021stable}. The tuning on the testing scene is carried out for $100,00$ iterations. The confidence threshold parameter $\tau$ is set to 0.05. The predefined depth threshold ($\sigma$) for applying Eq.($\textcolor{red}{8}$) in the main paper is two-thirds the median depth of the scene.  